\documentclass[sigconf]{acmart}

\AtBeginDocument{%
  }

\usepackage{multirow,multicol}
\usepackage{colortbl}
\definecolor{cvprblue}{rgb}{0.21,0.49,0.74}

\usepackage{graphicx}  
\usepackage{booktabs}     
\usepackage{subcaption}    
\definecolor{lightblue}{RGB}{224,232,240}

\copyrightyear{2026}
\acmYear{2026}
\setcopyright{cc}
\setcctype{by}
\acmConference[MM '26] {Proceedings of the 34th ACM International Conference on Multimedia}{November 10--14, 2026}{Rio de Janeiro, Brazil.}
\acmBooktitle{Proceedings of the 34th ACM International Conference on Multimedia (MM '26), November 10--14, 2026, Rio de Janeiro, Brazil}
\acmISBN{979-8-4007-2213-4/2026/11}
\acmDOI{10.1145/3767308.3834997}
\usepackage{pifont}

\makeatletter
\newcommand{\correspondingauthornote}{%
  \g@addto@macro\@authornotes{%
    \begingroup
      \renewcommand{\thefootnote}{\ding{41}}%
      \footnotetext{Corresponding author.}%
    \endgroup}}
\makeatother
\begin{document}

\title{Decoupled Similarity for Task-Aware Token Pruning in Large Vision-Language Models}

\author{Kexin Ma}
\orcid{0009-0000-8063-7316}
\email{kexin\_ma@whu.edu.cn}
\affiliation{%
  \department{School of Artificial Intelligence}
  \institution{Wuhan University}
  \city{Wuhan}
  \country{China}}

\author{Jing Xiao}
\orcid{0000-0002-0833-5679}
\email{jing@whu.edu.cn}
\affiliation{%
  \department{School of Artificial Intelligence}
  \institution{Wuhan University}
  \city{Wuhan}
  \country{China}}

\author{Chaofeng Chen}
\orcid{0000-0001-6137-5162}
\email{chaofengchen@whu.edu.cn}
\affiliation{%
  \department{School of Artificial Intelligence}
  \institution{Wuhan University}
  \city{Wuhan}
  \country{China}}

\author{Geyong Min}
\orcid{0000-0003-1395-7314}
\email{g.min@exeter.ac.uk}
\affiliation{%
  \department{Department of Computer Science}
  \institution{University of Exeter}
  \city{Exeter}
  \country{United Kingdom}}

\author{Guibo Zhu}
\orcid{0000-0001-8293-3952}
\author{Jinqiao Wang}
\orcid{0000-0002-9118-2780}
\affiliation{%
  \institution{Chinese Academy of Sciences}
  \city{Beijing}
  \country{China}}
\affiliation{%
  \institution{Wuhan AI Research}
  \city{Wuhan}
  \country{China}}

\author{Liang Liao}
\correspondingauthor
\correspondingauthornote
\orcid{0000-0002-2238-2420}
\email{liaoliang01@xidian.edu.cn}
\affiliation{%
  \department{Hangzhou Institute of Technology}
  \institution{Xidian University}
  \city{Hangzhou}
  \country{China}}

\begin{teaserfigure}
 \centering
  \includegraphics[width=0.98\textwidth]{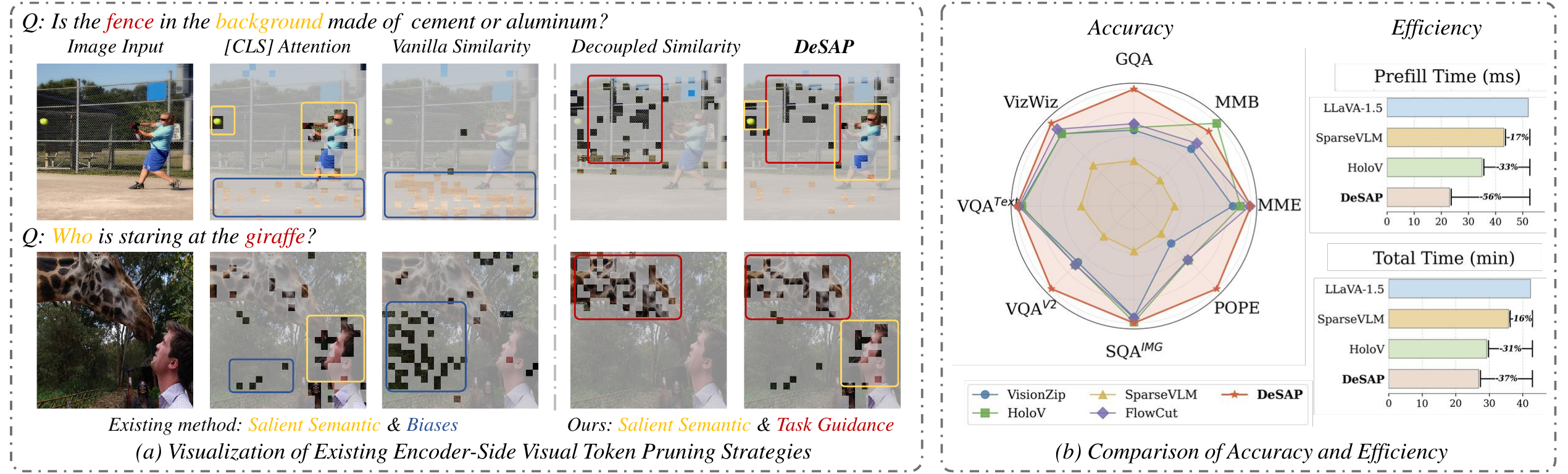}
    \caption{Advantages of the proposed DeSAP method for token pruning. (a) Visual comparisons on token pruning area: yellow - salient semantics, red - task-aware semantics, blue - bias.  (b) Quantitative comparison with existing SOTA token pruning methods on LLaVA-1.5-7B \cite{liu2024improved} at an 88.9\% pruning ratio. DeSAP consistently demonstrates superior accuracy and efficiency.} \label{fig0-1}
\end{teaserfigure}

\begin{abstract}
Token pruning has emerged as an effective approach to reduce the substantial computational overhead of Large Vision-Language Models (LVLMs) by discarding less informative visual tokens while preserving performance. However, existing methods typically rely on individual attention sources from different LVLM components, resulting in incomplete and suboptimal pruning decisions due to biased attention distributions. To address this, we propose DeSAP, a novel Decoupled Similarity-Aware Pruning method for precise, task-aware visual token pruning. Specifically, DeSAP introduces a decoupled similarity to capture fine-grained cross-modal relevance between visual and text tokens, providing explicit task-related guidance for pruning. By integrating this similarity with visual saliency signals derived from visual attention, DeSAP performs token pruning under the guidance of both task-related and visual cues, enabling robust pruning even under aggressive pruning ratios. Extensive experiments across diverse benchmarks show that DeSAP consistently outperforms existing methods in both accuracy and efficiency. 
\end{abstract}



\maketitle

\section{Introduction}
Large Vision-Language Models (LVLMs) have achieved remarkable progress \cite{liu2024improved,bai2025qwen25vltechnicalreport} in bridging visual and linguistic modalities, facilitating a wide range of downstream tasks such as visual question answering \cite{guo2023images, kuang2025natural}, multimodal reasoning \cite{li2025perception}, and video understanding \cite{lin2023video, wang2025internvideo}. However, these advances come at the cost of processing a large number of visual tokens, which often far exceeds the number of text tokens, especially for high-resolution images \cite{li2024mini} and multi-frame videos \cite{tang2025video}, leading to substantial computational overhead and hindering the efficiency and scalability of LVLMs in practical applications.

To address this issue, recent studies have proposed training-free token pruning methods that discard less informative tokens to reduce inference cost while maintaining performance \cite{wen2025tokenPI, yang2025visionzip, zhang2024sparsevlm}. Existing methods typically fall into two paradigms: (i) \textit{visual-centric pruning} \cite{bolya2022token, arif2025hired, shang2025llava} uses visual semantic information to retain visually salient or distinctive tokens, often derived from [CLS] attention in the visual encoder; (ii) \textit{task-centric pruning} \cite{chen2024image, zhang2024sparsevlm, xing2024pyramiddrop} utilizes cross-modal attention from LLMs to retain tokens relevant to the task. Despite their effectiveness, these methods typically rely on individual attention sources from different LVLM components, implicitly assuming that attention distributions serve as reliable proxies for token importance. However, recent studies suggest that attention distributions are often biased and may not faithfully reflect semantic importance. In particular, visual encoder attention tends to emphasize salient semantics, potentially overlooking fine-grained local details \cite{Registers, CalibCLIP,xu2024boosting}, while cross-modal attention may be biased toward specific spatial regions \cite{wen2025tokenPI, zhangvscan}. As a result, both paradigms suffer from incomplete or unreliable token selection, limiting their ability to preserve critical information under aggressive pruning.

Building on these limitations, a key observation is that existing pruning paradigms either lack task awareness (visual-centric) or introduce it too late (task-centric). This raises a natural question: \textit{can task-related information be incorporated earlier into encoder-side pruning ?} This could mitigate the global semantic bias of visual-centric pruning through introducing the task guidance, while avoiding the potential loss of visual cues in LLM-side pruning. To investigate this, we conduct an in-depth empirical study on existing pruning paradigms and explore how task guidance can be integrated into encoder-side token pruning. We obtain two key insights: (1) The global semantic bias in visual encoders not only resides in the [CLS] attention but also corrupts vanilla token-wise similarity estimation. This bias intensifies when the global semantics are weakly aligned with the text, resulting in persistent visual-textual misalignment and unreliable task-relevant importance estimation.
(2) By decomposing feature maps into residual and attention components, decoupled attention without residual connections preserves local details more effectively and captures finer-grained semantics. It also exhibits stronger alignment with text tokens, making it a better foundation for task-guided pruning before LLM decoding.

Based on the above observations, we propose DeSAP, a Decoupled Similarity-Aware Pruning method that enables fine-grained, task-guided visual token pruning before LLM decoding. DeSAP decomposes the final-layer visual feature into residual and attention components to decouple local details from global semantic proxies, and leverages fine-grained cross-modal alignment between the attention and text tokens to derive decoupled similarity as a reliable task cue. Combined with visual salient semantic cues, DeSAP performs token pruning under the joint guidance of both visual and task-related signals, mitigating representational collapse from both sides while preserving key information, enabling robust token pruning even under aggressive pruning ratios. The main contributions of this work are summarized as follows: 
\begin{itemize}
\item We present an in-depth empirical analysis of existing token pruning paradigms and introduce a novel perspective on visual encoder-side pruning by decoupling the visual features into residual and attention components, revealing decoupled attention's potential for preserving local details.

\item We propose DeSAP, a Decoupled Similarity-Aware Pruning method that leverages decoupled attention to derive precise task cues via fine-grained cross-modal alignment with text tokens, and integrates them with visual cues to enable robust token pruning before LLM decoding.

\item Extensive experiments on 8 benchmarks show that DeSAP consistently outperforms SOTA methods, achieving a 9$\times$ reduction in FLOPs and a 2.3$\times$ prefill speedup on LLaVA-1.5-7B while retaining only 11.1\% of visual tokens and maintaining 98.1\% of the original performance.
\end{itemize}

\section{Related Work}
\subsection{Efficiency in LVLMs}
Building on the capabilities of Large Language Models (LLMs) \cite{chung2024scaling, achiam2023gpt}, recent LVLMs have demonstrated remarkable performance in multimodal understanding \cite{guo2023images, kuang2025natural,wu2024q,Sun2026PromptImageCaptionCF} and reasoning \cite{li2025perception}. However, incorporating visual inputs introduces a large number of visual tokens, often exceeding text tokens and leading to substantial computational overhead \cite{vaswani2017attention, brauwers2021general}. These visual tokens are highly redundant with low information density, especially for high-resolution images. For instance, LLaVA-1.5 \cite{liu2024improved} encodes an image of resolution $336\times336$ into 576 visual tokens, while LLaVA-NeXT \cite{liu2024llava} and mini-Gemini-HD \cite{li2024mini} further increase the token count to 2,880. To alleviate this burden, several LVLMs introduce efficiency-oriented architectural designs, such as the Q-Former in InstructBLIP \cite{dai2023instructblip} and compression modules LLaMA-VID \cite{li2024llama} and DeCo \cite{yao2024deco}. Nevertheless, these approaches typically rely on additional training, which increases computational costs and limits scalability. 
\subsection{Vision Token Pruning in LVLMs}
Recent studies have explored training-free visual token pruning to enhance the efficiency of LVLMs, which can be broadly categorized into two paradigms:
(a) visual-centric strategies operate on the visual encoder, analyzing token distributions to discard visually redundant or semantically irrelevant ones. Early works reduce token redundancy through token merging \cite{wang2025folder, wen2025stop}. Later methods introduce importance-based selection that retains salient tokens while merging others to better preserve visual semantics. VisionZip \cite{yang2025visionzip} and LLaVA-PruMerge \cite{shang2025llava} employ [CLS] attention from the final encoder layer, whereas FlowCut \cite{tong2025flowcut} and HoloV \cite{zou2025don} further incorporate progressive pruning strategies to improve selection.
(b) Task-centric strategies perform token selection during the prefill stage of LLMs, leveraging cross-modal interactions to identify task-relevant tokens. SparseVLM \cite{zhang2024sparsevlm} adopts progressive cross-modal attention for task-guided pruning. PyramidDrop \cite{xing2024pyramiddrop} uses [EOS] attention to discard visual tokens. V$^2$Drop\cite{chen2025variation} further optimizes spatial bias in cross-modal attention. These methods typically rely on attention from individual LVLM components, with their effectiveness ultimately hinges on the distribution of attention.
\begin{figure}[t]
\centering
\includegraphics[width=0.98\columnwidth]{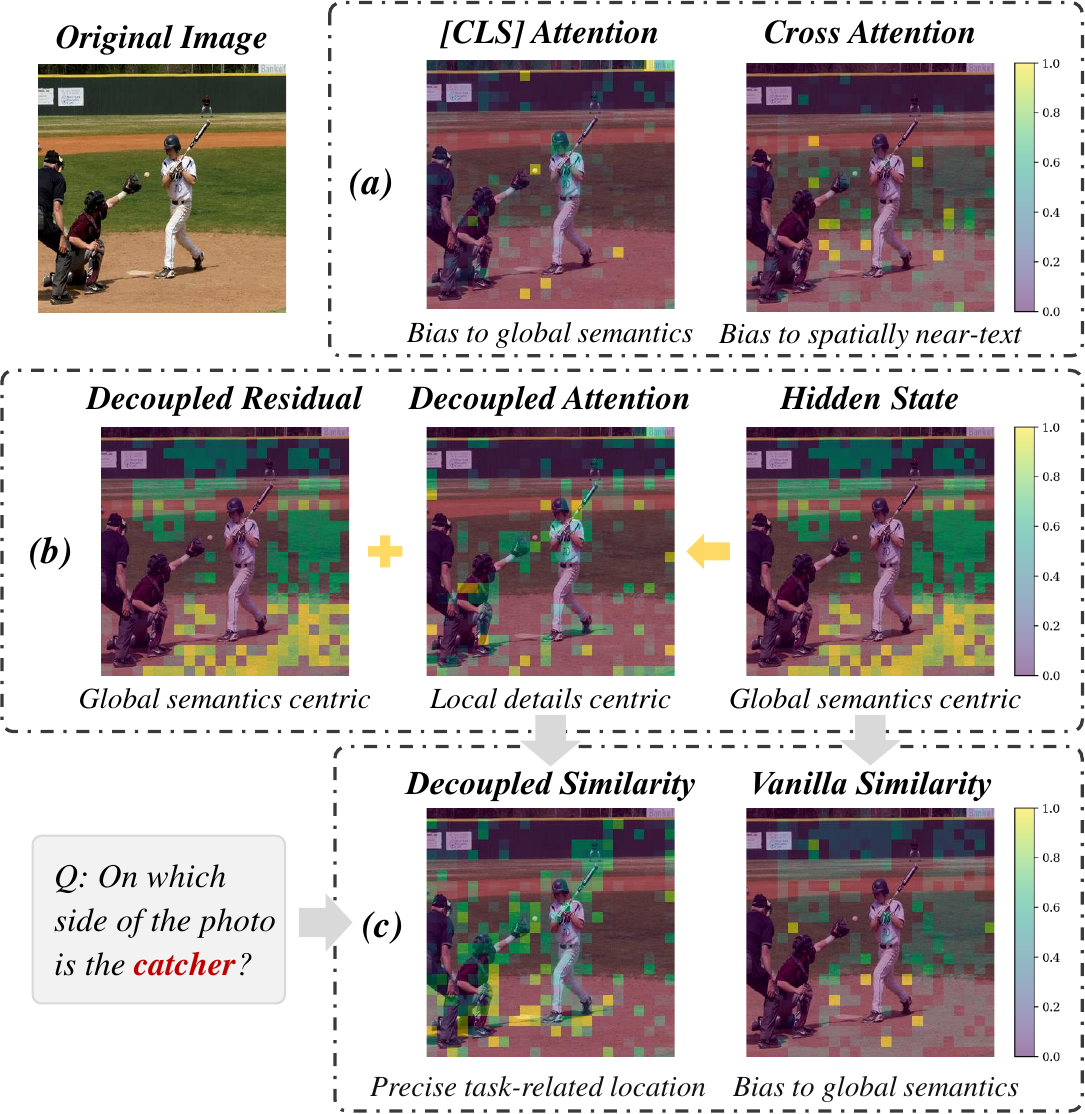}
\caption{Attention analysis in LVLMs. (a) [CLS] attention and cross-attention for pruning are biased toward global semantics and text-adjacent regions, respectively. (b) We separate detail-preserving attention from global semantic residuals by decomposing the hidden state. (c) Vanilla similarity is dominated by global semantics, while our decoupled similarity accurately identifies task-relevant local regions.} \label{fig1}
\end{figure}
\section{Empirical Analysis}
\subsection{Preliminary}
\textbf{Architecture of LVLMs.} 
We consider an LVLM parameterized by $\theta$, consisting of a visual encoder, a vision--text projector, a text embedding layer, and an LLM decoder. Given an image $v$ and a textual query $q$, the visual encoder and projector produce visual tokens $S_v$, while the query $q$ is mapped to text embeddings $S_q$. Their concatenation $[S_v,S_q]$ is fed into the LLM for autoregressive generation, with $y_t\sim p_\theta(y_t\mid S_v,S_q,y_{<t})$. 

\textbf{Vision Transformer (ViT).} 
The visual encoder in LVLMs is typically a Vision Transformer (ViT) \cite{DosovitskiyB0WZ21}. Its $l$-th residual block takes $S_v^{l-1}=[s_{\mathrm{cls}},s_1,\ldots,s_N]$ and applies multi-head self-attention followed by a feed-forward network, both with residual connections. We use $L$ to denote the final encoder layer, $proj(\cdot)$ the output projection, and $ffn(\cdot)$ and $ln(\cdot)$ the feed-forward network and layer normalization, respectively.

\begin{figure*}[h]
\centering
\includegraphics[width=\textwidth]{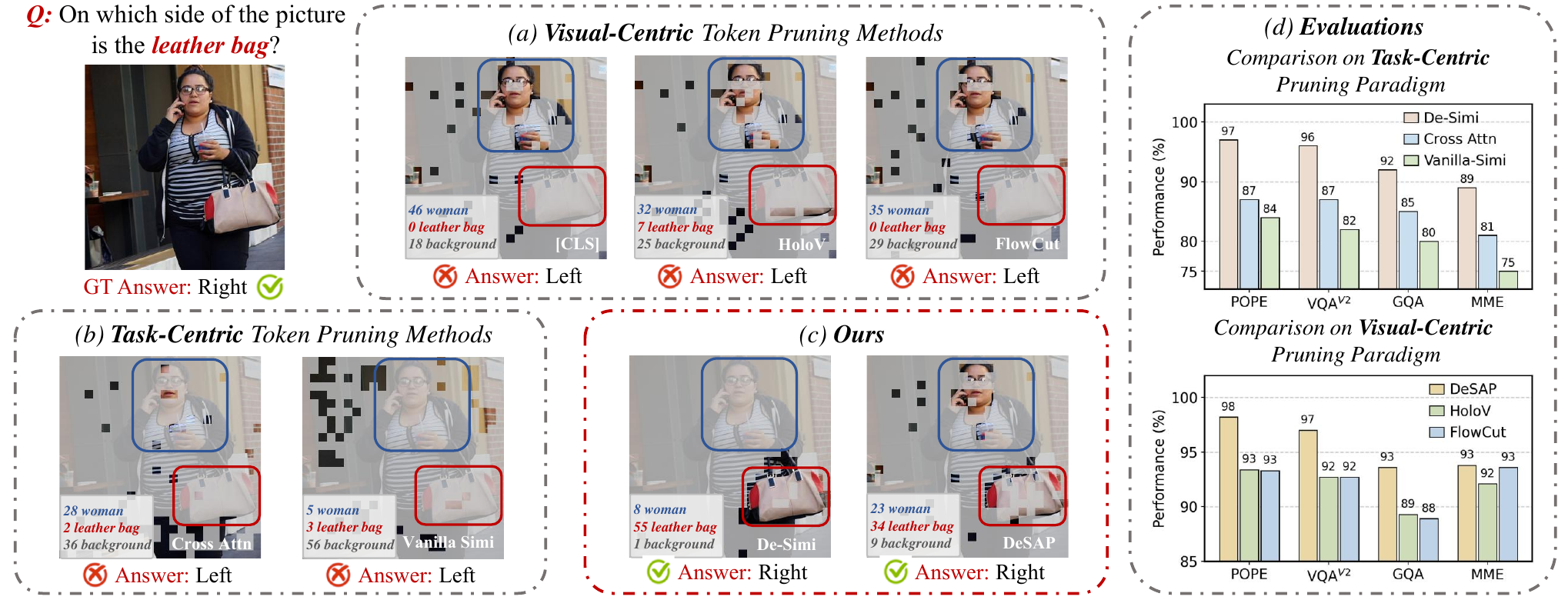}
\caption{Decoupled Similarity for Token Pruning. (a) Global bias in visual-centric pruning methods and limitations of two variant strategies. (b) Current task-centric pruning methods fail to leverage textual guidance. (c) Our decoupled similarity better localizes task-related regions and effectively exploits dual information for token pruning. (d) Evaluation results validate the effectiveness of decoupled similarity and its ability to integrate visual cues for dual-source guided token pruning.} \label{fig2}
\end{figure*}
\subsection{Attention Analysis in LVLMs}
Existing pruning methods commonly use attention to estimate token importance. We therefore analyze the attention signals used by both visual- and task-centric paradigms. For visualization, we average embeddings over features and attention over heads, normalize the scores to $[0,1]$, and reshape them into the patch grid.

\textbf{Attention Biases in LVLMs.} 
Visual-centric pruning methods typically retain tokens with high attention scores from the output of the visual encoder, such as [CLS] attention \cite{yang2025visionzip}. However, as shown in Fig.~\ref{fig1}(a), the [CLS] attention consistently focuses on broad contextual and salient regions rather than fine-grained local details. This suggests that the visual encoder is optimized to capture global semantics aligned with text, but may provide limited guidance for semantic details as well as fine-grained task-relevant localization.

Task-centric pruning attempts to identify token importance by leveraging cross-modal attention, starting from the early layers of the LLMs \cite{zhang2024sparsevlm, xing2024pyramiddrop}. However, as shown in Fig. \ref{fig1}(a), cross-attention in the early layers of LLMs exhibits a distinct spatial bias towards the image bottom. This bias arises because the instruction token used as the attention proxy favors nearby tokens \cite{vaswani2017attention}, with flattened bottom-image tokens positioned closest to it in the input sequence. Consequently, bottom regions are overemphasized, causing premature loss of useful visual information.

\textbf{Decomposition of Hidden State.} Motivated by ClearCLIP \cite{lan2024clearclip}, which shows that self-attention outputs disentangled from residual connections can better preserve fine-grained visual semantics, we decompose the hidden state ${S}_v^l$ from the last layer of the visual encoder, into two components: a decoupled residual component ${S}_{\text{d}} = {S}_v^{l-1}$, corresponding to the previous-layer hidden state, and a decoupled attention component $A_d = proj(A_l)$, corresponding to the projected self-attention output at layer $l$. Formally, the last-layer hidden state can be written as:
\begin{equation}
S_v^l = S_{\text{d}} + A_d + ffn\big(\ln(S_{\text{d}} + A_d\big).
\end{equation}
As illustrated in Fig.~\ref{fig1}(b), the decoupled residual component $S_d$ is nearly identical to the hidden state $\mathbf{S}_v^l$ and predominantly captures global semantic patterns concentrated in contextual background regions. In contrast, the decoupled attention component $A_d$ retains more localized and fine-grained semantic details, enabling a more distinctive focus on object regions such as the batter and catcher.

\textbf{Fine-grained Cross-modal Alignment via Decomposition.}
Recent methods \cite{chen2025recoverable, song2024less} estimate cross-modal alignment via cosine similarity between final-layer hidden states and text embeddings, as illustrated in Fig.~\ref{fig1}(c). However, we find that such similarity is predominantly governed by hidden-state activations, thereby inheriting their global semantic bias and consistently emphasizing contextual regions rather than fine-grained text-relevant details. Consequently, the vanilla similarity based on hidden states is insufficient for fine-grained cross-modal alignment, particularly when global semantics are weakly aligned with the text.

In contrast, we compute the cosine similarity between the decoupled attention $A_d$ and the text embeddings $S_q$. As shown in Fig. \ref{fig1}(c), this decoupled similarity more effectively highlights text-relevant regions under the guidance of detail-preserving attention, providing a more reliable foundation for fine-grained cross-modal alignment before LLM decoding.

\subsection{Decoupled Similarity for Token Pruning} 
Based on the above observations, decoupled similarity provides a more reliable and fine-grained task-guided signal for encoder-side token pruning. We further validate this through an empirical analysis of existing token pruning paradigms.

\textbf{Analysis of Limitations in Token Pruning Paradigms.} 
To better understand the behavior of existing token pruning methods, we visualize the tokens retained by representative approaches from two paradigms. (1) For visual-centric paradigm, we examine pruning methods based on [CLS] attention \cite{yang2025visionzip} and two variants HoloV \cite{zou2025don} and FlowCut \cite{tong2025flowcut} that attempt to mitigate attention bias through crop-wise or layer-wise adjustment. As shown in Fig. \ref{fig2}(a), pruning guided by [CLS] attention is heavily influenced by its global semantic bias, and both variants show limited performance in focusing on a wider range of semantics, including task-relevant regions. We attribute this limitation to the fact that the information sources remain confined to the encoder's attention, which also leads to complex reallocation strategies with non-negligible overhead. (2) For task-centric paradigm, we visualize task-guided pruning methods based on early cross-modal attention \cite{zhang2024sparsevlm} and vanilla similarity \cite{song2024less}. As shown in Fig. \ref{fig2}(b), they both exhibit noticeable bias, either toward bottom spatial regions or global semantics, and fail to reliably identify text-relevant regions. 

In contrast, the proposed decoupled similarity accurately highlights regions aligned with textual semantics, as shown in Fig.~\ref{fig2}(c). Furthermore, when combined with [CLS] attention under equal budgets, the resulting pruning efficiently leverages both visual saliency and task-specific signals. Visual attention provides coarse spatial guidance, while task-aware similarity refines the selection to precise regions, leading to more accurate and robust token selection.

\textbf{Quantitative Comparison with Token Pruning Paradigms.} To further support our analysis, we conducted two complementary experiments under the same pruning ratio (88.9\%) on LLaVA-1.5-7B \cite{liu2024improved}. First, we evaluate the ability of decoupled similarity to identify task-relevant regions by comparing it with cross-modal attention and vanilla similarity as importance measures for token pruning. As shown in Fig.~\ref{fig2}(d) \textit{(top)}, decoupled similarity consistently outperforms these methods, demonstrating more accurate identification of task-relevant tokens for pruning. We further compare our approach with methods designed to mitigate attention bias, including HoloV \cite{zou2025don} and FlowCut \cite{tong2025flowcut}. As illustrated in Fig.~\ref{fig2}(d) \textit{(bottom)}, our approach, which combines decoupled similarity with salient attention under equal budgets, achieves superior performance over these methods. This result highlights that jointly leveraging task and visual guidance is more effective than relying on bias correction within a single attention source.

\begin{figure*}[h]
\centering
\includegraphics[width=\textwidth]{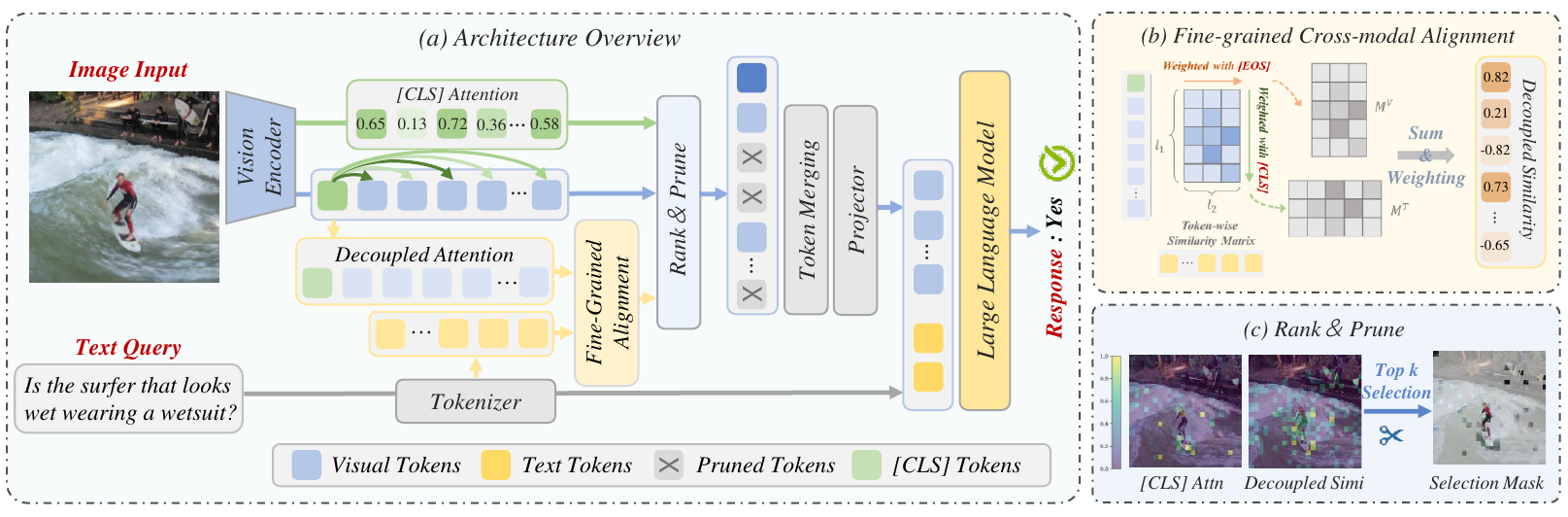}
\caption{Overall Architecture of DeSAP, a novel Decoupled Similarity-Aware Pruning method.  DeSAP obtains task guidance through fine-grained alignment between decoupled attention and text tokens, and combines it with salient semantic attention under equal token budgets for dual-source pruning. Discarded tokens are averaged into their most similar retained tokens.} \label{fig0}
\end{figure*}

\section{Methodology}
Based on the above analysis, we identify decoupled similarity as a reliable task-aware signal and propose DeSAP, a novel Decoupled Similarity-Aware Pruning method that integrates visual and textual semantic cues for pruning before LLM decoding. As illustrated in Fig. \ref{fig0}, by leveraging text guidance before LLM decoding, DeSAP enables dual-source guided token selection, effectively mitigating attention bias and improving pruning accuracy and efficiency.

\subsection{Task-relevant Semantics Alignment} 
As demonstrated in the previous section, decoupled attention remains sensitive to fine-grained local semantics, providing a reliable foundation for fine-grained task guidance. To further enhance this capability, we explore several attention variants based on the second last layer hidden state $S_v^{l-1}$, including the standard query-key attention $A_{qk}$, query-query attention $A_{qq}$, and value-value attention $A_{vv}$. Empirically, we find that $A_{qq}$ consistently performs best. We therefore formulate our decoupled attention based on $A_{qq}$ as:
\begin{equation}
A_{d} = proj(softmax\left(\frac{Q_s{Q_s}^T}{\sqrt{d}}\right) \cdot V_s),
\end{equation}
where $Q_s=\text{proj}_q(\text{ln}(S_v^{l-1})), V_s = \text{proj}_v(\text{ln}(S_v^{l-1}))$, and $d$ denotes the per-head feature dimension.

\textbf{Fine-grained Cross-modal Alignment.}
Direct cosine-similarity alignment between decoupled attention and text embeddings provides coarse task-relevant localization but remains sensitive to non-semantic tokens in both modalities. We therefore weight token-pair similarities using each token's relevance to the global representation of the complementary modality \cite{zou2022tokenflow}.

Let $A_d=[a_i]_{i=1}^{d_1}\in\mathbb{R}^{d_1\times e}$ and $S_q=[s_j]_{j=1}^{d_2}\in\mathbb{R}^{d_2\times e}$, where $d_1$ and $d_2$ denote the numbers of visual and non-padded text tokens, respectively, and $e$ is the embedding dimension. After $\ell_2$ normalization, we define the token-wise similarity matrix as
$M=[m_{ij}]=A_dS_q^\top$, where $m_{ij}=\cos(a_i,s_j)$ measures the similarity between the $i$-th visual token and the $j$-th text token.
We further define token-level weights
$\alpha_i=\cos(a_i,s_{\mathrm{eos}})$ and
$\beta_j=\cos(s_j,a_{\mathrm{g}})$,
where $a_{\mathrm{g}}$ and $s_{\mathrm{eos}}$ denote the visual global token and textual [EOS] representations, respectively. The bidirectional weighted similarities are:
\begin{equation}
\begin{aligned}
M^V_{ij}
&=\alpha_i
\frac{\exp(\lambda\beta_jm_{ij})}
{d_1\sum_{k=1}^{d_2}\exp(\lambda\beta_km_{ik})},\\
M^T_{ij}
&=\beta_j
\frac{\exp(\lambda\alpha_im_{ij})}
{d_2\sum_{k=1}^{d_1}\exp(\lambda\alpha_km_{kj})},
\end{aligned}
\end{equation}
where $i\in\{1,\ldots,d_1\}$, $j\in\{1,\ldots,d_2\}$, and $\lambda=0.1$ controls interaction sharpness. Accordingly, $m_{ik}$ compares visual token $i$ with text token $k$, whereas $m_{kj}$ compares visual token $k$ with text token $j$.
Consequently, we compute
$M_{V,T}=(M^V+M^T)\odot M$,
where $\odot$ denotes element-wise multiplication, and obtain the task-relevant attention map $A_t\in\mathbb{R}^{d_1}$ by averaging $M_{V,T}$ over the text dimension.

\subsection{Salient Semantic Integration}
Relying solely on task-relevant similarity may overemphasize task-specific semantics, particularly when the text underspecifies the target or the task requires additional context. We therefore leverage visual attention to capture holistic cues and extract a salient semantic attention map $A_s$ from the final visual-encoder layer. For models with a [CLS] token, we leverage its attention over all visual tokens; for models without a [CLS] token, we construct the attention by averaging the self-attention. The per-head scores and their average are computed as follows:
\begin{equation}
A_h=\operatorname{Softmax}\left(\frac{Q {K^{\top}}}{\sqrt{d_k}}\right), \quad {\overline{A}}=\frac{1}{H} \sum_{h=1}^H A_h.
\end{equation}
We denote the global semantic attention map as $A_s = \overline{A} \in \mathbb{R}^{d_{1}}$,  and leverage it as another information source alongside the task-relevant attention map $A_t$ for joint token importance assessment.

\subsection{Dual Information Guided Token Pruning}
We integrate the task-relevant attention map $A_t$ and global semantic attention map $A_s$ to jointly assess token importance.

\textbf{Rank \& Prune.}
Given a $K$-token budget, we first select the top $k_1=\lfloor K/2\rfloor$ tokens according to $A_t$ from the full token set $\mathcal{V}$ to form $\mathcal{I}_t$. We then select the top $k_2=K-k_1$ tokens according to $A_s$ from $\mathcal{V}\setminus\mathcal{I}_t$ to form $\mathcal{I}_s$. The final set $\mathcal{I}=\mathcal{I}_t\cup\mathcal{I}_s$ contains $K$ tokens, combining task relevance with global visual semantics.

\textbf{Token Merging.}
To mitigate information loss from discarded tokens, we employ a clustering-based merging strategy. Specifically, the retained token set $S_r$ is treated as cluster centers. Each unselected token $S_u$ is assigned to the cluster of its most similar center based on cosine similarity. The representation for each cluster is then computed by average merging \cite{bolya2022token}, resulting in the compressed output $S'_r$. This process effectively redistributes the information from discarded tokens without increasing token count.
\begin{table*}[!t]
 \begin{center}
 \caption{\textbf{Performance comparison of LLaVA-1.5-7B \cite{liu2024improved} under different pruning ratios.} The best and second-best results are \textbf{bolded} and \underline{underlined}, respectively, and the last column shows the average accuracy relative to the upper bound.}\label{tab:t1}
\footnotesize{
 \resizebox{0.9\textwidth}{!}{
  \begin{tabular}{l|c c c c c c c c c}
    \hline
    \textbf{Methods} & \textbf{GQA} & \textbf{MMB}  & \textbf{MME} & \textbf{POPE} & \textbf{SQA}$^\mathrm{IMG}$& \textbf{VQA}$^\mathrm{V2}$ &\textbf{VQA}$^\mathrm{Text}$ &\textbf{VizWiz}& \textbf{Avg.} \\
    \hline
    \rowcolor{gray!10} \multicolumn{10}{c} {\textit{UpperBound, 576 Tokens(100\%)}} \\ 
    \color{gray!60}\textbf{LLaVA-1.5-7B} \color{gray!60}\textit{(CVPR'24)} & \color{gray!60}61.9 & \color{gray!60}64.7  & \color{gray!60}1862 & \color{gray!60}85.9 & \color{gray!60}69.5 & \color{gray!60}78.5 & \color{gray!60}58.2 & \color{gray!60}50.0 & \color{gray!60}100.0\% \\
    \hline
        \rowcolor{gray!10} \multicolumn{10}{c} {\textit{Retain 64 Tokens on average ($\downarrow$ 88.9\%)}} \\ 
    \textbf{LLaVA-PruMerge} \textit{(ICCV'25)} & 51.9 &55.3 &1549 &65.3& 68.1 &67.4& 54.0 &50.1 &88.2\% \\
    \textbf{VisionZip}  \textit{(CVPR'25)} & 55.1& 60.1 &1690 &77.0& 69.0& 72.4 & 55.5&52.8&94.3\% \\
    \textbf{SparseVLM}  \textit{(ICML'25)} & 52.7& 56.2& 1505 &75.1 &62.2 &68.2 &51.8 &50.1&88.2\% \\
    \textbf{PyramidDrop}  \textit{(CVPR'25)} &  47.5 &58.8 & 1561 &55.9 &69.2 &69.2 &50.6 &50.7 &86.6\% \\
    \textbf{V$^2$Drop}  \textit{(CVPR'26)} &  50.5 &55.2 & 1470 &75.1 & 68.9 & - & 51.8 & - &86.9\% \\
     \textbf{FlowCut}  \textit{(NeurIPS'25)} & \underline{55.6} & 60.8 &  \underline{1744}  & 80.2  &69.1 &\underline{72.8}  & \underline{55.6} & \underline{53.2} &95.6\% \\
    \textbf{HoloV}  \textit{(NeurIPS'25)}  &  55.3 &\textbf{63.3} & 1715 & \underline{80.3}& \underline{69.5} & \underline{72.8} & 55.4 & 52.8 & \underline{95.8\%} \\    

    \rowcolor{lightblue!60} \textbf{DeSAP} \textit{(Ours)} & \textbf{58.3} & \underline{62.3} & \textbf{1748} & \textbf{85.8} & \textbf{69.6}  & \textbf{76.7} & \textbf{55.7} & \textbf{53.7} & \textbf{98.1}\% \\ 
    \hline
    \rowcolor{gray!10} \multicolumn{10}{c}{\textit{Retain 128 Tokens on average ($\downarrow$ 77.8\%)}}  \\
    \textbf{LLaVA-PruMerge}  \textit{(ICCV'25)} &53.3 &58.1 &1554 &67.2 &67.1& 68.8& 54.3 &50.3 &89.5\% \\
    \textbf{VisionZip}  \textit{(CVPR'25)} & 57.6 & 62.0 & 1763 & 83.2 & 68.9 & 75.6 & 56.8 & 51.6 & 97.1\% \\
    \textbf{SparseVLM}  \textit{(ICML'25)} & 56.0 & 60.0  & 1696 & 80.5 & 67.1 & 73.8 & 54.9 & 51.4 & 94.5\% \\
    \textbf{PyramidDrop}  \textit{(CVPR'25)} & 57.1 & 61.6 & 1761 & 82.3 & 68.4 & 72.9 & 56.6 & 51.0 & 96.0\% \\
    \textbf{V$^2$Drop}  \textit{(CVPR'26)} &  56.3 & 61.8 & 1712 &80.9 &68.8 &- & 53.8 &- &94.0\% \\
     \textbf{FlowCut}  \textit{(NeurIPS'25)}  &  \underline{58.5} &62.1 & 1792 & \underline{85.2} &68.6 & \underline{76.0} &\textbf{57.3} &\underline{52.2} &98.0\% \\
    \textbf{HoloV}  \textit{(NeurIPS'25)}  &  57.7 &\textbf{63.9} & \underline{1802} &84.0& \underline{69.8}& 75.5  &56.8& 51.5 &\underline{98.1\%} \\ 
    \rowcolor{lightblue!60} \textbf{DeSAP} \textit{(Ours)} & \textbf{59.5} & \underline{62.8}  & \textbf{1815} & \textbf{87.0} & \textbf{69.9} & \textbf{76.9} & \underline{57.0} & \textbf{54.3} & \textbf{99.6}\% \\ 
    \hline
\rowcolor{gray!10} \multicolumn{10}{c} {\textit{Retain 192 Tokens on average($\downarrow$ 66.7\%)}} \\ 
    \textbf{LLaVA-PruMerge}  \textit{(ICCV'25)} & 54.3 &59.6  &1632 &71.3 &67.9 &70.6 &54.3 &50.1 &91.5\% \\
    \textbf{VisionZip}  \textit{(CVPR'25)} & 59.3 & 63.0  & 1783 & 85.3 & 68.9 & \underline{77.4} & 57.3 & 51.2 & 98.4\% \\
    \textbf{SparseVLM}  \textit{(ICML'25)} & 57.6 & 62.5  & 1721 & 83.6 & 69.0 & 75.6 & 56.1 & 50.5 & 96.5\% \\
    \textbf{PyramidDrop}  \textit{(CVPR'25)} & 57.3 & 63.3  & 1797 & 82.3 & 69.0 & 75.1 & 56.5 & 51.1 & 97.1\% \\
    \textbf{V$^2$Drop}  \textit{(CVPR'26)} & 58.5 &63.7 & 1826 &85.1 &69.3 &- &55.6 &- &97.6\% \\
     \textbf{FlowCut}  \textit{(NeurIPS'25)}  &  \underline{60.1} & 63.2  &  \underline{1836} & \underline{86.1}& 68.6&  77.1  & \textbf{57.5} & \underline{51.8} &99.1\% \\
    \textbf{HoloV}  \textit{(NeurIPS'25)}  &  59.0 &\textbf{65.4} & 1820 &85.6& \underline{69.8} & 76.7&\underline{57.4}& 50.9 &\underline{99.0}\% \\ 

   \rowcolor{lightblue!60}  \textbf{DeSAP} \textit{(Ours)} & \textbf{60.3} & \underline{63.5}  & \textbf{1848}  & \textbf{87.1} & \textbf{69.9} & \textbf{77.5} & 57.2 & \textbf{54.5} & \textbf{100.3}\% \\ 
    \hline
  \end{tabular}}}
 \end{center}
\end{table*}
\begin{table*}[!t]
 \begin{center}
 \caption{\textbf{Performance comparison of LLaVA-NeXT-7B \cite{liu2024llava}.} The best and second-best results are \textbf{bolded} and \underline{underlined}, respectively, and the last column shows the average accuracy relative to the upper bound. }\label{tab:t2}
\footnotesize{
 \resizebox{0.9\textwidth}{!}{
  \begin{tabular}{l| c c c c c c c c c}
    \hline
    \textbf{Methods} & \textbf{GQA} & \textbf{MMB}  & \textbf{MME} & \textbf{POPE} & \textbf{SQA}$^\mathrm{IMG}$& \textbf{VQA}$^\mathrm{V2}$ &\textbf{VQA}$^\mathrm{Text}$ &\textbf{VizWiz}& \textbf{Avg.} \\
    \hline
    \rowcolor{gray!10} \multicolumn{10}{c} {\textit{Upper Bound, 2880 Tokens (100\%)}} \\ 
    \color{gray!60}\textbf{LLaVA-NeXT-7B} \color{gray!60}\textit{(CVPR'24)} & \color{gray!60}64.2 & \color{gray!60}67.4  & \color{gray!60}1851 & \color{gray!60}86.5 & \color{gray!60}70.1 & \color{gray!60}81.8  & \color{gray!60}61.3 & \color{gray!60}57.6  & \color{gray!60}100.0\% \\
    \hline
    \rowcolor{gray!10} \multicolumn{10}{c} {\textit{Retain 320 Tokens ($\downarrow$ 88.9\%)}} \\ 
    \textbf{SparseVLM}  \textit{(ICML'25)} & 56.1 & 60.6  &1533 & 82.4  &66.1 & 71.5 & 58.4 & 52.0 & 90.3\% \\
    \textbf{PyramidDrop}  \textit{(CVPR'25)} & 56.4 &63.4 &1663 &77.6& 67.5 &73.5& 54.4 &54.1& 91.3\% \\
    \textbf{LLaVA-PruMerge}  \textit{(ICCV'25)} & 53.6 &61.3& 1534 &60.8& 66.4& 69.7 &50.6 &54.0 & 85.5\% \\
    \textbf{VisionZip}  \textit{(CVPR'25)} & 59.3& 63.1 &1702& 82.1& 67.3& 76.2 &\underline{58.9}&-& 94.0\% \\
    \textbf{Hired}  \textit{(AAAI'25)} & 59.3 &64.2& 1690& 83.3& 66.7& 75.7 &58.8& 54.2& 94.1\% \\
    \textbf{FlowCut} \textit{(NeurIPS'25)}  & 59.8& 65.3 &\underline{1791}& 83.4 & -  & 77.8 & 60.1 &-&96.1\% \\
    \textbf{HoloV}  \textit{(NeurIPS'25)} & \underline{61.7} &\textbf{65.3}   & 1738  &\underline{83.9} & \underline{68.9}  &\underline{79.5}  &58.7  &\underline{55.3}  &96.4\%\\   
   \rowcolor{lightblue!60}  \textbf{DeSAP} \textit{(Ours)} & \textbf{62.5} & \underline{64.8}  & \textbf{1781} & \textbf{87.5} & \textbf{69.3} & \textbf{80.6} & \textbf{59.3} & \textbf{57.3} & \textbf{98.0}\% \\ 
    \hline
  \end{tabular}}}
 \end{center}
\end{table*}

\begin{figure*}[t]
\centering
\includegraphics[width=0.96\textwidth]{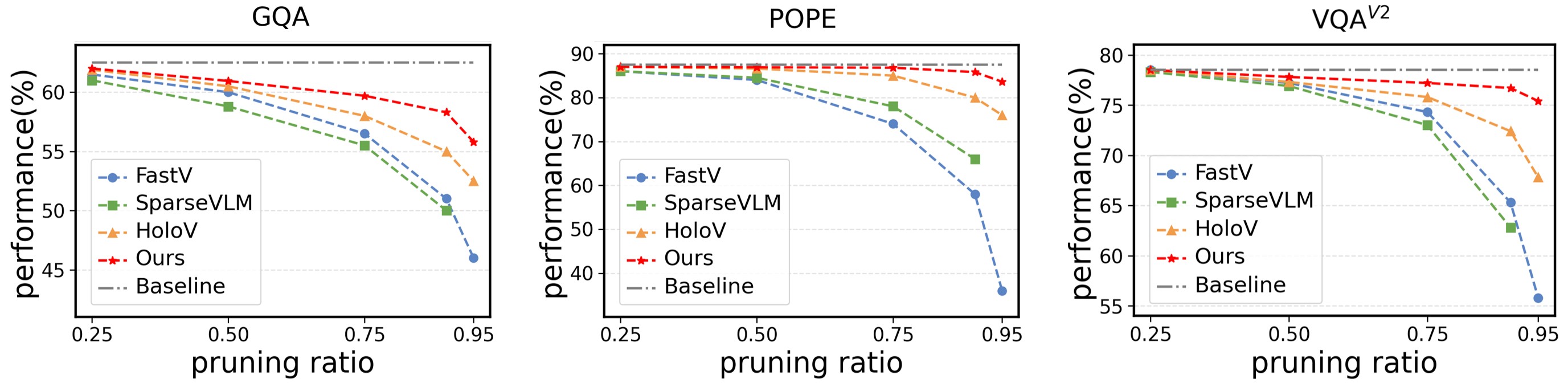}
\caption{Performance comparison of various methods on LLaVA-1.5-7B across multiple benchmarks at different pruning ratios.} \label{fig_exp_plot}
\end{figure*}
\section{Experiments}
\subsection{Experimental settings}
\textbf{Benchmarks.} We evaluate the effectiveness of our method on 8 standard visual understanding benchmarks, including visual question answering benchmarks such as GQA \cite{hudson2019gqa}, ScienceQA \cite{lu2022learn}, VQAv2 \cite{goyal2017making}, TextVQA \cite{singh2019towards} and VizWiz \cite{gurari2018vizwiz}; multi-modal reasoning benchmarks such as MMBench \cite{liu2024mmbench}, MME \cite{chaoyou2023mme} and POPE \cite{li2023evaluating}.

\textbf{Models and Baselines.} To assess the generalizability of DeSAP, we evaluate it on three representative large vision-language models including LLaVA-1.5-7B \cite{liu2024improved}, LLaVA-NeXT-7B \cite{liu2024llava}, and Qwen-2.5-VL-7B \cite{bai2025qwen25vltechnicalreport}, and we compare its performance with 9 state-of-the-art token pruning methods, including visual centric methods such as LLaVA-PruMerge \cite{shang2025llava}, VisionZip \cite{yang2025visionzip}, FlowCut \cite{tong2025flowcut}, HoloV \cite{zou2025don}, HiRED \cite{arif2025hired}, FastV \cite{chen2024image}, and task-centric methods including SparseVLM \cite{zhang2024sparsevlm}, PyramidDrop \cite{xing2024pyramiddrop}, V$^2$Drop \cite{chen2025variation}. 

\subsection{Results and Discussions}
\textbf{Results on LLaVA-1.5-7B.}
As shown in Table \ref{tab:t1}, we evaluate the performance of DeSAP on LLaVA-1.5-7B \cite{liu2024improved} against state-of-the-art methods over 8 image understanding benchmarks. With token budgets of 64, 128, and 192 per image, DeSAP maintains performance close to the full model, with marginal accuracy drops of only 1.9\% at 64 tokens and 0.4\% at 128 tokens. Remarkably, at 192 tokens, it surpasses the full-model performance by 0.3\%. Moreover, DeSAP excels on tasks with a strong semantic focus, such as GQA \cite{hudson2019gqa}, POPE \cite{li2023evaluating}, VizWiz \cite{gurari2018vizwiz} and VQAv2 \cite{goyal2017making}. As shown in Fig. \ref{fig_exp_plot}, DeSAP consistently outperforms SOTA approaches across pruning ratios from 0.25 to 0.95, achieving the best performance on all four benchmarks. Furthermore, DeSAP shows a clear advantage under high pruning ratios. With an aggressive 88.9\% token reduction rate, it outperforms all other methods on 7 out of 8 benchmarks, and surpasses the second-best method, HoloV \cite{zou2025don}, by 2.3\% on average.

\begin{figure*}[t]
\centering
\includegraphics[width=\textwidth]{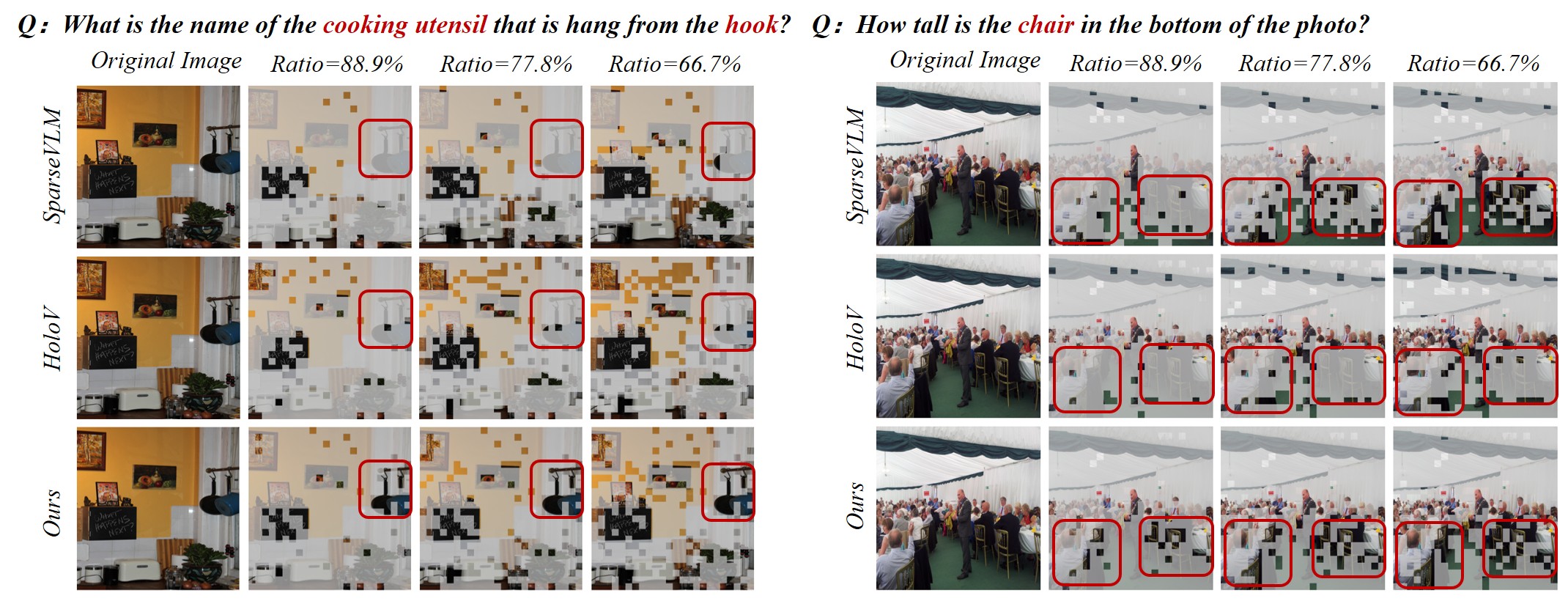}
\caption{Visualization on GQA comparing SparseVLM, HoloV, and Ours.
The original image and the corresponding pruned results are shown at pruning ratios of 88.9\%, 77.8\%, and 66.7\%. Gray areas indicate discarded tokens, while red regions highlight the primary task-relevant semantic tokens. Compared with other methods, our method well preserves critical tokens.} \label{fig_exp}
\end{figure*}
\begin{table}[t]
 \begin{center}
 \caption{\textbf{Comparative Experiments on Qwen2.5-VL-7B \cite{bai2025qwen25vltechnicalreport}.} The best results are \textbf{bolded}.}
 \label{tab:t3} 
 \resizebox{\linewidth}{!}{
  \begin{tabular}{l|c c c c c c}
    \toprule
    \textbf{Methods} & \textbf{GQA} & \textbf{MME} & \textbf{POPE} & \textbf{SQA}  & \textbf{VQA}$^\mathrm{Text}$& \textbf{Avg.} \\
    \hline
    \rowcolor{gray!10} \multicolumn{7}{c} {\textit{Upper Bound  (100\%)}} \\ 
    \color{gray!60}\textbf{Qwen2.5-VL-7B}  &  \color{gray!60} 60.6   & \color{gray!60}2304 & \color{gray!60}86.1 & \color{gray!60}84.7 & \color{gray!60}84.8 & \color{gray!60}100.0\% \\
    \hline
    
\rowcolor{gray!10} \multicolumn{7}{c} {\textit{Token Pruning Rate = 88.9\% $\downarrow$ }} \\ 
    \textbf{FastV} \textit{(ECCV'24)}  & - & 1940 &78.6& 77.4 & 60.3& 84.5\% \\
    \textbf{HoloV} \textit{(NeurIPS'25)}  & -  &2006 & 80.7 & 79.5 & 61.8 & 86.9\% \\

\rowcolor{lightblue!60} \textbf{DeSAP} \textit{(Ours)} & \textbf{57.5} & \textbf{2056}  & \textbf{83.5} & \textbf{80.3}  & \textbf{63.2} & \textbf{90.1\%} \\ 

       \hline
\rowcolor{gray!10} \multicolumn{7}{c} {\textit{ Token Pruning Rate = 77.8\% $\downarrow$ }} \\ 
    \textbf{FastV} \textit{(ECCV'24)}  & - & 2036 &80.7 &78.0 & 69.0 & 88.9\% \\
    \textbf{HoloV} \textit{(NeurIPS'25)}  & - & 2043& 82.3 &79.8 &70.3 & 90.3\% \\

\rowcolor{lightblue!60} \textbf{DeSAP} \textit{(Ours)} & \textbf{58.9}  & \textbf{2110}  & \textbf{85.7} & \textbf{80.5} & \textbf{72.1}  & \textbf{93.7\%} \\ 
   \hline
   
    \rowcolor{gray!10} \multicolumn{7}{c} {\textit{ Token Pruning Rate = 66.7\% $\downarrow$ }} \\ 
    \textbf{FastV} \textit{(ECCV'24)}  & - &2072 & 82.2 & 78.5 & 77.9 &92.5\% \\
    \textbf{HoloV} \textit{(NeurIPS'25)}  & - &2093 & 85.0 & 79.8 & 78.9 & 94.2\% \\

    \rowcolor{lightblue!60} \textbf{DeSAP} \textit{(Ours)} & \textbf{59.8} & \textbf{2183} & \textbf{86.5} & \textbf{80.6}  & \textbf{79.5}  & \textbf{96.6\%} \\

   \bottomrule
  \end{tabular}} 
 \end{center}
\end{table}

\textbf{Results on Qwen-2.5-VL-7B.}
To further validate the generalizability of DeSAP beyond the LLaVA-based frameworks, we performed experiments on Qwen2.5-VL-7B \cite{bai2025qwen25vltechnicalreport}. As shown in Table \ref{tab:t3}, we find that DeSAP consistently outperforms the SOTA method HoloV \cite{zou2025don} across various reduction ratios. It achieves average performance of 90.1\%, 93.7\% and 96.6\% at 88.9\%, 77.8\% and 66.7\% pruning ratios, surpassing HoloV \cite{zou2025don} by 3.2\%, 3.4\% and 2.4\%, respectively. These results demonstrate the robustness and adaptability of DeSAP across diverse model architectures, ensuring stable performance even under aggressive pruning ratios.

\textbf{Results with Higher Resolution.}
We further evaluate DeSAP on the advanced LLaVA-NeXT-7B \cite{liu2024llava} across different benchmarks mentioned above, with comparison to current SOTA approaches. As shown in Table \ref{tab:t2}, under a fixed token budget of 320 per image, equivalent to an 88.9\% reduction, our method consistently delivers superior performance, achieving an average of 98.0\% of full-model performance. It ranks first on 7 out of 8 benchmarks and surpasses the second-best method, HoloV \cite{zou2025don}, by an average of 1.6\%.

\begin{table}[t]
    \caption{Ablation experiment on LLaVA-1.5-7B \cite{liu2024improved} at a pruning ratio of 88.9\%, analyzing (a) Retention rates $T$ and $S$ for task-related and salient semantic attention; (b) Decoupled attention encoding strategy $E$, with all other settings fixed.}  \label{tab:t4}
    \begin{subtable}{0.53\linewidth}
      \centering
        \caption{Retention rates $T$ and $S$.}
        \resizebox{\linewidth}{!}{
        \begin{tabular}{ccccc}
            \toprule
     $\textbf{T}$ & $\textbf{S}$ & \textbf{GQA} &  \textbf{POPE} & \textbf{MME} \\
     \hline
     0\% & 100\%& 57.2  & 82.8 & 1664 \\ 
     25\% &75\% & 57.8  & 85.3  & 1728 \\ 
     \rowcolor{lightblue!60} 50\% &50\%  &\textbf{58.3} &\textbf{85.8} &\textbf{1748}\\ 
     75\%  &25\%  & 58.2   &85.8   & 1740 \\ 
     100\% &0\%   & 58.1  &85.6   & 1686  \\ 
            \bottomrule
        \end{tabular}
    }
    \end{subtable}
    \begin{subtable}{0.45\linewidth}
      \centering
        \caption{Encoding strategy.}
        \resizebox{\linewidth}{!}{
        \begin{tabular}{l c c c}
            \toprule
    $\textbf{E}$ & \textbf{GQA} & \textbf{POPE} & \textbf{MME} \\
     \hline 
     $qkv$ &58.1   &85.2 & 1741\\ 
     \rowcolor{lightblue!60} $qqv$ &\textbf{58.3}   &\textbf{85.9}  & \textbf{1748}\\ 
     $vvv$ &58.2   &85.8  & 1697\\ 
     $kkv$ &58.3   &85.5  & 1720\\ 

     \bottomrule
        \end{tabular}
        }
    \end{subtable} 
\end{table}

\subsection{Ablation Studies}
\textbf{Varying Retention Ratios \textit{T} and \textit{S}.}
In Table \ref{tab:t4}(a), we investigate the trade-off in allocating pruning ratios between task-relevant \textit{T} and salient semantics attention \textit{S}. The results show that an equal allocation between the two leads to optimal performance. On datasets with semantically clear instructions, such as GQA and POPE, assigning a larger proportion of the budget to task-relevant semantics results in only a slight performance drop. However, on datasets where global semantics play a more dominant role, such as MME, integrating global semantic attention remains indispensable.

\textbf{Decoupled Attention Encoding Strategy \textit{E}.}
We compare different decoupled attention encoding strategies as discussed in Section 4.1. As shown in Table \ref{tab:t4}(b), $xyz$ denotes $softmax(XY^\top/\sqrt{d})Z$; e.g., $qqv$ uses $QQ^\top$ to aggregate $V$. As shown in Table~\ref{tab:t4}(b), all formulations perform comparably, while $qqv$ (i.e., $A_{qq}$) achieves the best and most stable results and is therefore adopted in our formulation.

\textbf{Ablation on Fine-grained Cross-modal Alignment.}
To further investigate the effect of fine-grained cross-modal alignment strategy (FCMA), we conduct a controlled comparison on LLaVA-1.5-7B \cite{liu2024improved} under an 88.9\% pruning ratio with all other variables fixed. As shown in Table \ref{tab:t5}, we consider four variants, including vanilla similarity, vanilla similarity with FCMA, our decoupled similarity, and our decoupled similarity with FCMA. The results indicate that FCMA performs similarly or worse when applied to vanilla similarity, which already performs poorly for token pruning. In contrast, our decoupled similarity alone achieves strong results with an average accuracy of 94\%, and integrating it with FCMA further boosts performance by 0.9\%.

\begin{table}[!t]
 \begin{center}
 \caption{\textbf{Ablation on fine-grained cross-modal alignment.} All variants retain 64 tokens under identical settings.} \label{tab:t5}
\scriptsize{
 \resizebox{\linewidth}{!}{
  \begin{tabular}{l| l l l l l l }
    \hline
    \textbf{Methods} & \textbf{GQA} & \textbf{MME} & \textbf{POPE} & \textbf{SQA} & \textbf{VQA}$^\mathrm{T}$ &\textbf{Avg.} \\
    \hline
    \rowcolor{gray!10} \multicolumn{7}{c} {\textit{Upper Bound, 576 Tokens (100\%)}} \\ 
    \color{gray!60}\textbf{LLaVA-1.5-7B}  & \color{gray!60}61.9  & \color{gray!60}1862& \color{gray!60}85.9 & \color{gray!60}69.5 & \color{gray!60}58.2&  \color{gray!60}100.0\%  \\
    \hline
    \rowcolor{gray!10} \multicolumn{7}{c} {\textit{Retain 64 Tokens ($\downarrow$ 88.9\%)}} \\ 

    \textbf{Vanilla-Simi}  & 50.2  & 1355 & 72.1   & 66.5    & 46.2  & 81.8\% \\
    + alignment & 50.4 & 1362  & 72.5 &65.7& 45.4& 81.7\%\\
    
    \hline
   \textbf{De-Simi} & 58.0 & 1667 & 85.2  &68.5  &52.3 &94.0\% \\
  \rowcolor{lightblue!60}  + alignment & \textbf{58.3} & \textbf{1686}& \textbf{85.7}&\textbf{68.9}& \textbf{53.5}& \textbf{94.9}\%\\

    \hline
    
  \end{tabular}}
  }
 \end{center}
\end{table}

\subsection{Visualization Analysis}
For qualitative comparison, we visualize the retained visual patches under different pruning ratios of 88.9\%, 77.8\%, and 66.7\% on GQA \cite{hudson2019gqa}. As shown in Figure \ref{fig_exp}, the gray areas represent discarded tokens, while the red regions highlight the primary semantic areas relevant to the task. Compared to SparseVLM \cite{zhang2024sparsevlm} and HoloV \cite{zou2025don}, DeSAP preserves more task-related and contextual visual cues, even under a high pruning ratio (\emph{e.g.}, 88.9\%). It effectively filters out redundant visual tokens while retaining crucial ones, further demonstrating the effectiveness of our framework.

\begin{table}
 \begin{center}
 \caption{\textbf{Efficiency comparisons on the POPE benchmark.}\label{tab:t6}
 We report the practical total running time (min:sec), prefilling time, TFLOPs and the achieved accuracy.}
 \resizebox{\linewidth}{!}{
  \begin{tabular}{l|c c c c}
    \toprule
    \textbf{Methods}  & \textbf{Total Time} $\downarrow$  & \textbf{Prefill Time}  $\downarrow$  & \textbf{FLOPs}  $\downarrow$  & \textbf{Acc.} $\uparrow$  \\ 
    \hline
    \color{gray!60}LLaVA-1.5-7B  &\color{gray!60}42:20 \footnotesize{(1.0×)} & \color{gray!60}52 \color{gray!60}ms \footnotesize{(1.0×)} &\color{gray!60}3.82 T  &\color{gray!60}85.9   \\
    \textbf{+SparseVLM}    &  35:38 \footnotesize{(1.2×)} & 43 ms \footnotesize{(1.2×)} & 1.31 T &  75.1   \\
    \textbf{+HoloV}   &  29:12 \footnotesize{(1.4×)} & 35 ms \footnotesize{(1.5×)} & 0.42 T &  80.3 \\
   \rowcolor{lightblue!60}   \textbf{+Ours}  &  \textbf{ 26:35 \footnotesize{(1.6×)}} &  \textbf{ 22 ms \footnotesize{(2.3×)}} & \textbf{0.42}  T & \textbf{85.8}  \\
   \bottomrule
  \end{tabular}}
 \end{center}
\end{table}

\subsection{Efficiency Analysis}
As shown in Table \ref{tab:t6}, we assess the practical acceleration achieved by DeSAP on the POPE benchmark \cite{li2023evaluating}. By retaining only 11.1\% of visual tokens, DeSAP achieves a 1.6$\times$ overall speedup, a 2.3$\times$ prefill speedup, and a 9.1$\times$ reduction in FLOPs on LLaVA-1.5-7B \cite{liu2024improved}, with only a 0.1\% accuracy drop. Compared with SparseVLM \cite{zhang2024sparsevlm}, DeSAP reduces FLOPs from 1.31T to 0.42T while improving accuracy by 10.7\%. Moreover, under the same FLOPs as HoloV \cite{zou2025don}, DeSAP improves accuracy by 5.5\% and reduces prefill latency from 35 to 22 ms, demonstrating a superior efficiency-accuracy trade-off.

\section{Conclusion}
In this work, we provide a detailed analysis of existing token pruning paradigms and their limitations. Based on this analysis, we propose DeSAP, achieving fine-grained cross-modal alignment before LLM decoding while leveraging visual guidance for context preservation. Its dual-source design preserves complementary task-relevant and contextual cues, enabling robust selection under aggressive pruning. Extensive experiments show that DeSAP consistently outperforms current SOTA methods in both accuracy and efficiency, demonstrating the strong capabilities of our approach.

\section*{Acknowledgement}
This work was supported by the National Natural Science Foundation of China (62371351), the Natural Science Foundation of Hangzhou (2025SZRJJ2224), the Key R\&D Program of Hubei Province (2025BEB011), and the Intelligent Computing Center of the National Cybersecurity Talent and Innovation Base, Wuhan.

\bibliographystyle{ACM-Reference-Format}
\bibliography{sample-base}

\clearpage
\appendix

\section*{Appendix}

In the Appendix, we provide additional details and experimental results to enhance understanding and insights into our method. The Appendix is organized as follows:

Section \ref{sec:Exp} presents additional experimental results that further validate the effectiveness and robustness of our method; 

Section \ref{sec:details} provides extended experimental details, including FLOPs calculation and full experiment configurations, to facilitate reproducibility; 

Section \ref{limitation} discusses the limitations of this work and explores its broader implications and impacts.

\section{Additional Experimental Results} \label{sec:Exp}
\subsection{Results on Video-LLaVA}

\begin{table}[h]
 \vspace{-2mm}
 \begin{center}
 \caption{Performance comparisons on Video-LLaVA-7B \cite{lin2023video} across 2 video understanding tasks with a 50\% reduction rate.} \label{video}
 \resizebox{\linewidth}{!}{
  \begin{tabular}{l|c c c c c c }
      \toprule
    \multirow{2}{*}{\textbf{Methods}} & \multicolumn{2}{c}{\textbf{MSVD-QA}} & \multicolumn{2}{c}{\textbf{MSRVTT-QA}} & \multicolumn{2}{c}{\textbf{Average}} \\
                             & Acc.         & Score        & Acc.         & Score         & Acc.         & Score        \\ 
    \midrule
    \color{gray!60}\textbf{Video-LLaVA-7B}      & \color{gray!60}70.2 &\color{gray!60}3.9 &\color{gray!60}57.3& \color{gray!60}3.5 &\color{gray!60}63.8 & \color{gray!60}3.7  \\
    \textbf{FastV} \footnotesize\textit{ECCV'24}        & 71.0 &3.9 &55.0& 3.5  & 63.0 &3.7 \\
    \textbf{HoloV}   \footnotesize\textit{NeurIPS'25}     & 71.0 &\textbf{4.0} &56.5 &\textbf{3.6}& 63.7& \textbf{3.8}  \\
   \rowcolor{gray!10}  \textbf{Ours}    &  \textbf{71.2} & \textbf{4.0} & \textbf{57.0} &\textbf{3.6} & \textbf{64.1} &\textbf{3.8}  \\ 
   \bottomrule
  \end{tabular}}
 \end{center}
 \vspace{-2mm}
\end{table}

To comprehensively evaluate the effectiveness of our approach, we conduct experiments on video understanding tasks using Video-LLaVA-7B \cite{lin2023video} and compare it with current SOTA methods HoloV \cite{zou2025don} and FastV \cite{chen2024image}. As shown in Table \ref{video}, our method consistently maintains competitive performance across multiple video understanding benchmarks while retaining only 50\% of the visual token budget. Notably, it outperforms all compared methods and preserves nearly 100\% of the original model performance, or even surpasses it on MSVD-QA \cite{chen2011collecting} benchmarks.

\subsection{Qualitative Results}
\begin{figure*}[t]
\centering
\includegraphics[width=0.98\textwidth]{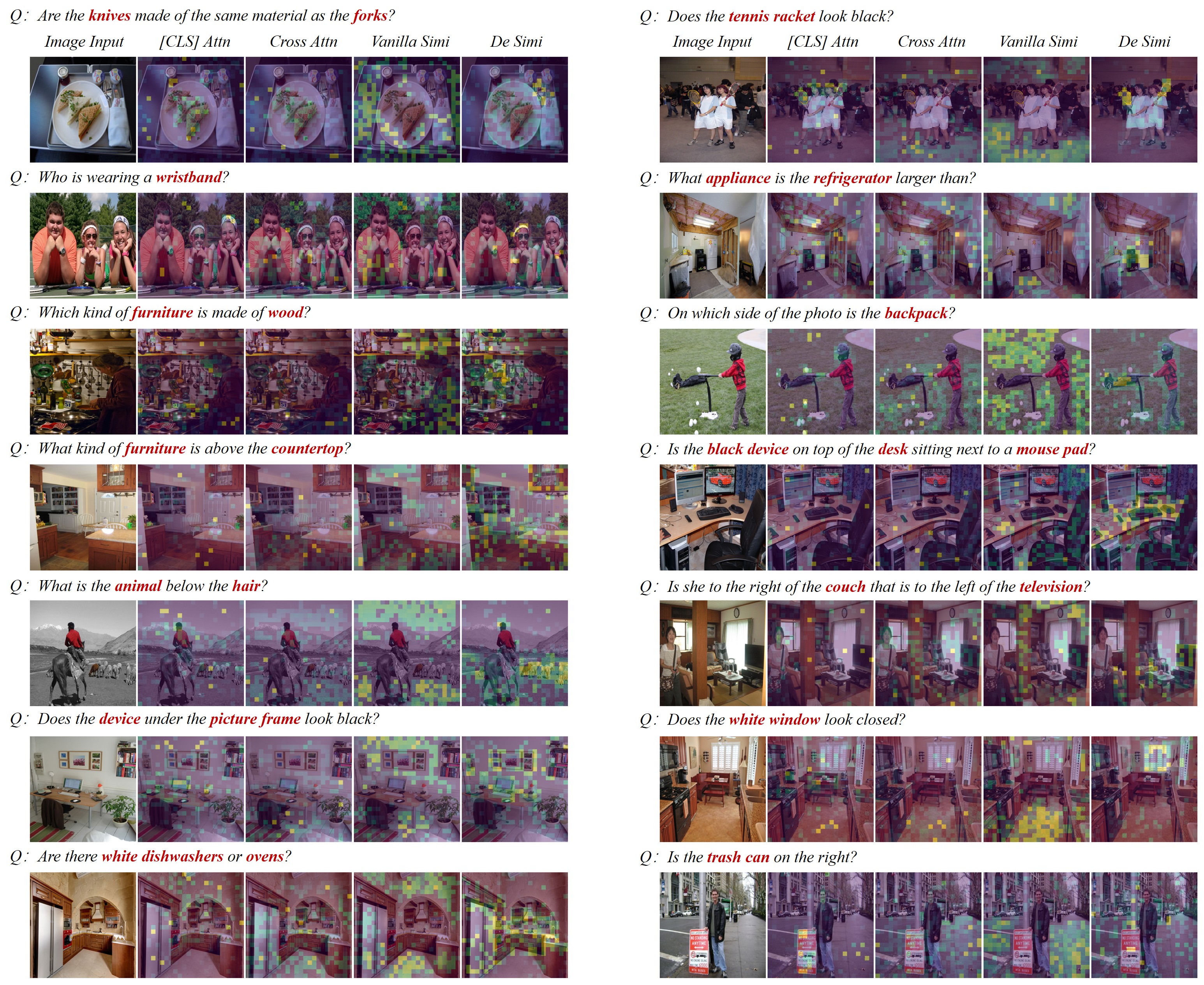}
\caption{Qualitative comparison of activation maps from [CLS] attention, cross-attention, vanilla similarity, and our decoupled similarity. While the first three exhibit noticeable biases, \textit{our method consistently achieves precise localization aligned with the text-query semantics}, demonstrating its effectiveness for encoder-side semantic alignment in token pruning.} \label{sup_vis_0}
\end{figure*}

To further validate the effectiveness of our method, we provide additional qualitative comparisons of [CLS] attention, cross-attention, vanilla similarity, and our proposed decoupled similarity. As shown in Figure \ref{sup_vis_0}, [CLS] attention consistently emphasizes global and salient semantics, whereas cross-attention tends to activate the bottom image regions. Vanilla similarity is continuously biased by global semantic. In contrast, \textit{our decoupled similarity more precisely localizes the text-referred semantic region across different queries.} These results further support the effectiveness of our method and highlight its contribution, while also suggesting its potential to benefit future research on encoder-side semantic alignment.

Moreover, We provide a detailed visualization of the retained visual patches under different pruning rates 88.9\%, 77.78\% and 66.7\% on LLaVA-1.5-7B \cite{liu2024improved} with GQA \cite{hudson2019gqa} and POPE \cite{li2023evaluating} benchmarks, to further demonstrate the effectiveness of our proposed DeSAP. Specifically, we compare DeSAP with three SOTA pruning methods: FlowCut \cite{tong2025flowcut}, HoloV \cite{zou2025don}, and SparseVLM \cite{zhang2024sparsevlm}. FlowCut \cite{tong2025flowcut} and HoloV \cite{zou2025don} represent the line of methods that perform token selection based on [CLS] attention, whereas SparseVLM \cite{zhang2024sparsevlm} represents methods that rely on cross-modal attention for token selection.

\begin{figure*}[t]
\centering
\includegraphics[width=0.76\textwidth]{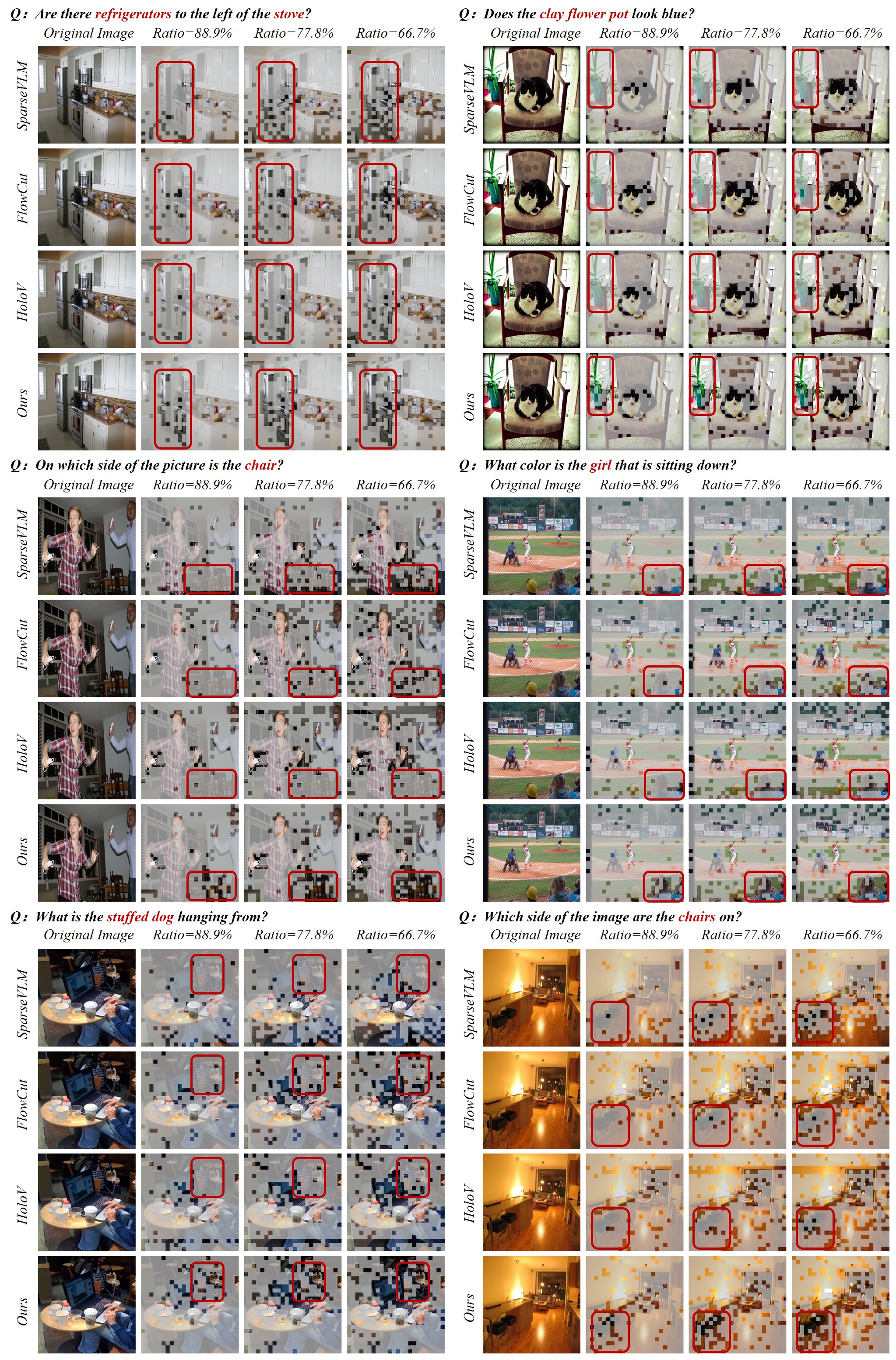}
\caption{\textit{Qualitative comparison of SparseVLM \cite{zhang2024sparsevlm}, HoloV \cite{zou2025don}, FlowCut \cite{tong2025flowcut}, and our proposed method DeSAP on the GQA dataset \cite{hudson2019gqa}.} The figure presents original images alongside their pruned versions at pruning ratios of 88.9\%, 77.8\%, and 66.7\%. Bounding boxes are used to highlight key semantic regions aligned with the text.  Our method demonstrates superior preservation of key semantics, especially under high pruning ratios (e.g., 88.9\%).} \label{sup_vis_1}
\end{figure*}

\begin{figure*}[t]
\centering
\includegraphics[width=0.76\textwidth]{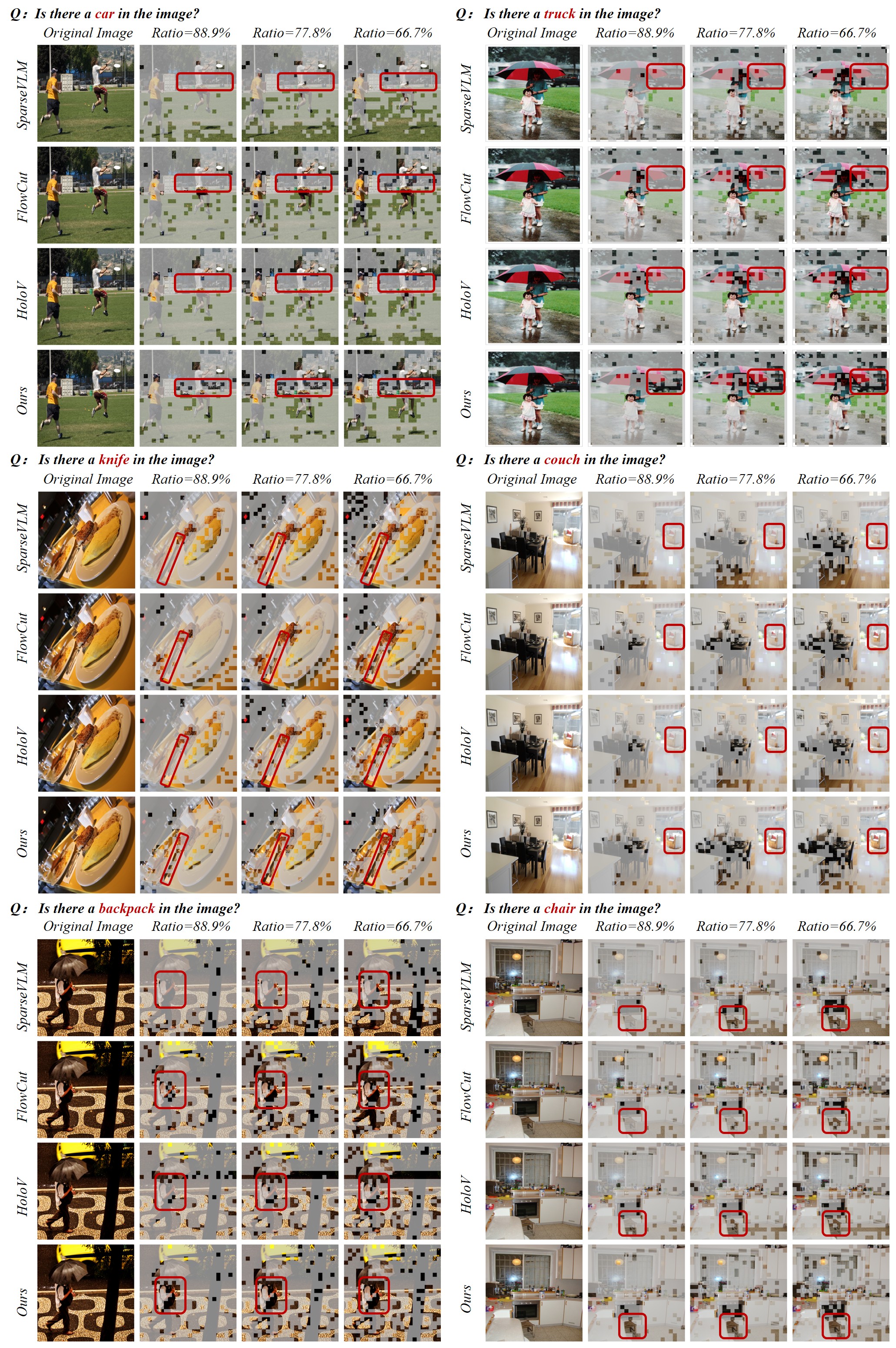}
\caption{\textit{Qualitative comparison of SparseVLM \cite{zhang2024sparsevlm}, HoloV \cite{zou2025don}, FlowCut \cite{tong2025flowcut}, and our proposed method DeSAP on the POPE dataset \cite{li2023evaluating}.} The figure presents original images alongside their pruned versions at pruning ratios of 88.9\%, 77.8\%, and 66.7\%. Bounding boxes are used to highlight key semantic regions aligned with the text.  Our method demonstrates superior preservation of key semantics, especially under high pruning ratios (e.g., 88.9\%).} \label{sup_vis_2}
\end{figure*}

As shown in Figures \ref{sup_vis_1} and \ref{sup_vis_2}, the gray regions denote discarded tokens, while the red bounding boxes highlight the key semantic regions aligned with the task. Compared with existing state-of-the-art methods, our approach consistently preserves more task-relevant visual cues while retaining richer global semantic information. As a result, it is capable of responding to a broader range of open-ended questions while maintaining stable performance even under aggressive pruning ratios (e.g., 88.9\%). Moreover, the figures show that our method is effective not only for large objects (Figure \ref{sup_vis_1}), but especially for small objects occupying only a limited portion of the image (Figure \ref{sup_vis_2}). \textit{For such small-object scenarios, both salient-semantic and global-semantic approaches often struggle to capture the relevant regions, whereas our method can precisely identify them according to the textual query.}

\section{More Experimental Details} \label{sec:details}

\subsection{Implementation Details}
We follow the standard inference configurations provided in the official implementations of each evaluated LVLMs. 
All of our experiments are conducted on Nvidia H20-96G GPU. The implementation was carried out in Python 3.10.18, utilizing PyTorch 2.1.2, and CUDA 12.1. All baseline settings follow the original paper. We adhere to the standard evaluation protocol established by LLaVA \cite{liu2024improved}. For both LLaVA-1.5-7B \cite{liu2024improved} and Qwen-VL-2.5 \cite{bai2025qwen25vltechnicalreport}, we follow the respective models' original inference configurations as provided in their public repositories. Similarly, for video understanding tasks, we adopt the inference setup specified in the official Video-LLaVA \cite{tang2025video} codebase. Moreover, for datasets containing a large number of text tokens as questions, such as MMBench \cite{liu2024mmbench} and ScienceQA \cite{lu2022learn}, we employ text token importance assessment as proposed by CalibCLIP \cite{CalibCLIP} to filter out numerous words irrelevant to the key semantics and only uses key semantic text tokens for similarity assessment, thereby reducing unnecessary computational overhead.

\subsection{Computational Complexity}
To evaluate the computational complexity of LVLMs, we analyze their core components, including the self-attention mechanism and Feed-Forward Network (FFN). Following prior work such as PyramidDrop\cite{xing2024pyramiddrop} and FastV\cite{chen2024image}, we estimate the theoretical floating-point operations (FLOPs) introduced in the LLM decoding layers during the pre-filling stage for processing the visual tokens. The total FLOPs can be expressed as:
\begin{equation}
\text{Total FLOPs}=\sum_{k=1}^K\left(4 n_k d^2+2 n_k^2 d+3 n_k d m\right), 
\end{equation}
where $K$ denotes the number of transformer layers, $n_k$ is the number of visual tokens at layer $k$, $d$ is the hidden dimension size, and  $m$ is the intermediate dimension of FFN. This formulation highlights the substantial impact of visual token sequence length $n$ on overall computational load. After token pruning, the FLOPs can be reformulated as:
\begin{equation}
\text{Post-Pruning FLOPs}=\sum_{k=1}^K\left(4 \hat{n}_k d^2+2 \hat{n}_k^2 d+3 \hat{n}_k d m\right),
\end{equation}
where $\hat{n}_k$ refers to the number of tokens retained at layer $k$ after pruning. The theoretical reduction ratio of FLOPs attributed to visual token pruning is then computed as:
\begin{equation}
R = 1-\frac{\text {Post-Pruning FLOPs}}{\text {Total FLOPs}}.
\end{equation}
This analysis demonstrates that reducing the number of visual tokens can greatly lower the computational cost during inference.

\textit{Notably, the additional overhead of our fine-grained cross modal alignment is minimal.} The additional cost of the proposed fine-grained cross-modal alignment is mainly dominated by the matrix multiplications for computing $m_s$, $\alpha_s$, and $\beta_t$, resulting in a complexity of $\mathcal{O}(d_1 d_2 e)$, where $d_1$, $d_2$, and $e$ denote the numbers of visual tokens, text tokens, and the embedding dimension, respectively. In practice, since the text length $d_2$ and embedding dimension $e$ are typically fixed and much smaller than the number of visual tokens, the computational overhead scales approximately linearly with respect to the number of visual tokens.

Specifically, for a representative setting with $A_d \in \mathbb{R}^{576 \times 768}$ and $S_q \in \mathbb{R}^{30 \times 768}$, \textit{the main computation requires approximately $2.7 \times 10^7$ FLOPs per sample}, excluding the minor overhead of element-wise normalization. \textit{This cost is negligible compared with the reduced backbone computation of 3.4 TFLOPs.}

\subsection{Backbones and Baselines}
\textbf{Models.} To assess the general applicability and robustness of our approach, we experiment with multiple widely adopted large vision-language models (LVLMs) featuring diverse architectures. Specifically, we follow common practice in this research domain and conduct comparisons using LLaVA-1.5-7B \cite{liu2024improved}, a widely used academic model that converts each input image into 576 visual tokens. We also include LLaVA-NeXT-7B \cite{liu2024llava}, which further enhances high-resolution image comprehension by representing an input with up to 2,880 tokens. For video understanding tasks, we employ Video-LLaVA-7B \cite{lin2023video}, extending the framework to process up to 8 video frames represented by 2,048 visual tokens. Additionally, we provide some of the first experimental evaluations on the recently released Qwen-2.5-VL \cite{bai2025qwen25vltechnicalreport}, which supports processing images with a variable number of tokens ranging from 4 to 16,384.

\textbf{Baselines.} We compare our method against several state-of-the-art methods for visual token reduction.
(1) FastV \cite{chen2024image}, which prunes visual tokens at early LLM layers according to text-oriented attention weights.
(2) SparseVLM \cite{zhang2024sparsevlm}, which estimates visual token importance by analyzing vision-relevant text tokens and incorporates adaptive sparsity ratios with token recycling.
(3) PyramidDrop \cite{xing2024pyramiddrop}, which drops a fraction of visual tokens progressively after each model stage, forming a pyramidal token structure.
(4) V$^2$Drop \cite{chen2025variation}, which aims to alleviate the spatial bias in LLM cross-attention, thereby enabling more precise task-guided token pruning.
(5) LLaVA-PruMerge \cite{shang2025llava}, which combines pruning of low-importance tokens via sparse [CLS] attention and merges retained tokens via key clustering.
(6) VisionZip \cite{yang2025visionzip}, which selects dominant tokens and performs contextual merging of the remaining tokens.
(7) FlowCut \cite{tong2025flowcut}, which mitigates the global-semantic bias of [CLS]-based pruning by progressively pruning tokens from shallow to deep encoder layers, rather than applying pruning only at the final layer.
(8) HoloV \cite{zou2025don}, which mitigates global bias in a spatially aware manner by defining crops at the final encoder layer and reallocating the pruning ratio across different crops.
(9) HiRED \cite{arif2025hired}, which dynamically allocates token budgets across image patches based on [CLS] attention and selects informative tokens per region.
These methods highlight diverse design choices for visual token pruning, \textit{ranging from [CLS]-attention-based strategies to cross-attention-based approaches, and collectively advance the acceleration of LVLMs.}

\subsection{Benchmarks and Metrics}
We perform comprehensive evaluations across a diverse set of standard vision-language benchmarks, encompassing both image and video understanding tasks.

\textbf{Visual question answering benchmark.} 
(1) GQA \cite{hudson2019gqa}: Constructed from scene graphs, questions, and images, this benchmark evaluates compositional reasoning and visual scene understanding through structured visual question answering.
(2) ScienceQA \cite{lu2022learn}: Features a hierarchically organized question set spanning natural, language, and social sciences, with 26 topics and 379 skills for evaluating multimodal reasoning and explanatory capabilities.
(3) VQAv2 \cite{goyal2017making}: Tests visual perception through open-ended questions about 265K real-world images, with human-annotated answers enabling robust evaluation of question-answering accuracy.
(4) TextVQA \cite{singh2019towards}: Assesses reading comprehension and visual-textual integration by requiring models to interpret text embedded within images to answer questions.
(5) VizWiz \cite{gurari2018vizwiz}: A visual question answering dataset built from images taken by blind people to support assistive technology development.

\textbf{Visual question answering benchmark.}
(1) MMBench \cite{liu2024mmbench}: A hierarchical evaluation framework that systematically assesses model capabilities across three levels, spanning from fundamental perception and reasoning (L-1) to 20 fine-grained ability dimensions (L-3).
(2) MME \cite{chaoyou2023mme}: A holistic benchmark comprising 14 specialized subtasks designed to measure diverse perceptual and cognitive abilities while minimizing data contamination risks through concise instruction-answer design.
(3) POPE \cite{li2023evaluating}: Specifically targets object hallucination assessment by presenting binary questions about object presence, evaluated through accuracy, precision, recall, and F1-score across multiple sampling strategies.

\textbf{Video Understanding Benchmarks.} We extend our evaluation to four video question-answering datasets:
(1) MSVD-QA \cite{chen2011collecting}: Comprises 1,970 video clips with approximately 50K QA pairs across five question types (what, who, how, when, where) for video captioning and QA evaluation.
(2) MSRVTT-QA \cite{xu2016msr}: A larger-scale video QA dataset containing 10K clips and 243K question-answer pairs, challenging models to integrate visual and temporal information for accurate response generation.

Following established practices in the field \cite{xing2024pyramiddrop}, we employ both accuracy and GPT-based evaluation scores as primary metrics for video understanding tasks, while using dataset-specific official metrics for image understanding benchmarks.

\section{Limitations and Broader Impacts} \label{limitation}
\textbf{Limitations.}
DeSAP is the first method to achieve fine-grained cross-modal alignment purely on the encoder side, enabling efficient and precise task-guided pruning by exploiting the substantially lower token dimension in the encoder compared with the LLM. Nevertheless, our method still has several limitations.

First, when the text query fails to explicitly highlight the key semantics, our method becomes more dependent on the [CLS]-based salient semantic signal. This limitation is especially noticeable in captioning-style tasks, where the text often provides weak or diffuse semantic guidance. A promising direction is to introduce a dynamic trade-off between salient-semantic and text-semantic pruning ratios, which requires further study. Second, long textual inputs increase the cost of similarity computation. To mitigate this issue, we employ the text token importance assessment strategy from CalibCLIP \cite{CalibCLIP} to remove tokens irrelevant to the key semantics and retain only informative text tokens for similarity estimation. However, designing a simpler and more efficient text token selection mechanism remains an important direction for future research. Despite these limitations, we believe DeSAP still offers valuable insights into general semantic task-guided pruning, especially in scenarios that do not require long text contexts.

\textbf{Boarder Impacts.}
The development of efficient Large Vision-Language Models (LVLMs) has the potential to benefit a broad range of domains, including autonomous systems, robotics, healthcare, education, and assistive technologies. By incorporating the DeSAP visual token pruning mechanism, LVLMs are able to remove a large proportion of task-irrelevant visual tokens directly within the visual encoder. This significantly reduces the computational cost imposed on downstream LLMs and broadens the deployment potential of these systems. Such efficiency is particularly valuable in compute-constrained settings, for example, onboard satellite platforms where visual data must be processed or transmitted to ground stations under limited resource budgets.

\end{document}